\documentclass[runningheads]{llncs}
\usepackage{graphicx}
\usepackage{tikz}
\usepackage{comment}
\usepackage{amsmath,amssymb} % define this before the line numbering.
\usepackage{color}

\usepackage{epsfig}
\usepackage{graphicx}
\usepackage{amsmath}
\usepackage{amssymb}
\usepackage{multirow}
\usepackage{enumerate}
\usepackage{bm}
\usepackage{wrapfig}
\usepackage{epstopdf}
\usepackage{wrapfig}
\usepackage{hyperref}

\DeclareMathOperator*{\argmin}{argmin}

\usepackage{cuted}
\usepackage{capt-of}

\begin{document}
% \renewcommand\thelinenumber{\color[rgb]{0.2,0.5,0.8}\normalfont\sffamily\scriptsize\arabic{linenumber}\color[rgb]{0,0,0}}
% \renewcommand\makeLineNumber {\hss\thelinenumber\ \hspace{6mm} \rlap{\hskip\textwidth\ \hspace{6.5mm}\thelinenumber}}
% \linenumbers
\pagestyle{headings}
\mainmatter
\def\ECCVSubNumber{4607}  % Insert your submission number here

\title{Latent Partition Implicit with Surface Codes for 3D Representation} % Replace with your title

% INITIAL SUBMISSION
\begin{comment}
\titlerunning{ECCV-22 submission ID \ECCVSubNumber}
\authorrunning{ECCV-22 submission ID \ECCVSubNumber}
\author{Anonymous ECCV submission}
\institute{Paper ID \ECCVSubNumber}
\end{comment}
%******************

% CAMERA READY SUBMISSION
%\begin{comment}
\titlerunning{Latent Partition Implicit with Surface Codes for 3D Representation}
% If the paper title is too long for the running head, you can set
% an abbreviated paper title here
%
%\author{Chao Chen\inst{1}\orcidID{0000-1111-2222-3333} \and
%Yu-Shen Liu\inst{1}\orcidID{1111-2222-3333-4444} \and
%Zhizhong Han\inst{2}\orcidID{2222--3333-4444-5555}}
\author{Chao Chen\inst{1} \and
Yu-Shen Liu\inst{1}\thanks{The corresponding author is Yu-Shen Liu. This work was supported by National Key R$\&$D Program of China (2020YFF0304100), the National Natural Science Foundation of China (62072268), and in part by Tsinghua-Kuaishou Institute of Future Media Data.} \and
Zhizhong Han\inst{2}}
\authorrunning{C. Chen et al.}
% First names are abbreviated in the running head.
% If there are more than two authors, 'et al.' is used.
%
\institute{School of Software, BNRist, Tsinghua University, Beijing, China \and
Department of Computer Science, Wayne State University, Detroit, USA\\
\email{chenchao19@mails.tsinghua.edu.cn, liuyushen@tsinghua.edu.cn, h312h@wayne.edu}}
%\institute{School of Software, BNRist, Tsinghua University, Beijing, China \and
%Springer Heidelberg, Tiergartenstr. 17, 69121 Heidelberg, Germany
%\email{lncs@springer.com}\\
%\url{http://www.springer.com/gp/computer-science/lncs} \and
%ABC Institute, Rupert-Karls-University Heidelberg, Heidelberg, Germany\\
%\email{\{abc,lncs\}@uni-heidelberg.de}}
%\end{comment}
%******************
\maketitle

\begin{abstract}
Deep implicit functions have shown remarkable shape modeling ability in various 3D computer vision tasks. One drawback is that it is hard for them to represent a 3D shape as multiple parts. Current solutions learn various primitives and blend the primitives directly in the spatial space, which still struggle to approximate the 3D shape accurately. To resolve this problem, we introduce a novel implicit representation to represent a single 3D shape as a set of parts in the latent space, towards both highly accurate and plausibly interpretable shape modeling. Our insight here is that both the part learning and the part blending can be conducted much easier in the latent space than in the spatial space. We name our method \textit{Latent Partition Implicit (LPI)}, because of its ability of casting the global shape modeling into multiple local part modeling, which partitions the global shape unity. LPI represents a shape as Signed Distance Functions (SDFs) using surface codes. Each surface code is a latent code representing a part whose center is on the surface, which enables us to flexibly employ intrinsic attributes of shapes or additional surface properties. Eventually, LPI can reconstruct both the shape and the parts on the shape, both of which are plausible meshes. LPI is a multi-level representation, which can partition a shape into different numbers of parts after training. LPI can be learned without ground truth signed distances, point normals or any supervision for part partition. LPI outperforms the latest methods under the widely used benchmarks in terms of reconstruction accuracy and modeling interpretability. Our code, data and models are available at \href{https://github.com/chenchao15/LPI}{https://github.com/chenchao15/LPI}.
%\dots
\end{abstract}

\section{Introduction}
Implicit functions have been a popular representation for 3D objects or scenes. They are able to describe the 3D geometry using the occupancy information~\cite{MeschederNetworks,chen2018implicit_decoder} or signed distances~\cite{DBLP:journals/corr/abs-1901-06802,Park_2019_CVPR} at arbitrary query locations. We can leverage deep neural networks to learn implicit functions, which we call deep implicit functions. Deep implicit functions usually regard input images or point clouds as conditions to discriminate different object identities in different applications, such as single image reconstruction~\cite{xu2019disn,pifuSHNMKL19,DBLP:conf/cvpr/ChibaneAP20,Gidi_2019_ICCV,Genova:2019:LST,seqxy2seqzeccv2020paper} or surface reconstruction~\cite{Williams_2019_CVPR,liu2020meshing,Mi_2020_CVPR,Genova:2019:LST}. Although deep implicit functions have the remarkable ability of geometry modeling, it is difficult for them to decompose shapes into parts. This significantly limits the part-based modeling or editing and makes geometry modeling not interpretable.

\begin{figure*}[tb]
  \centering
  % the following command controls the width of the embedded PS file
  % (relative to the width of the current column)
  %\includegraphics[width=.95\linewidth, bb=39 696 126 756]{figures/definition3.eps}
   \includegraphics[width=\linewidth]{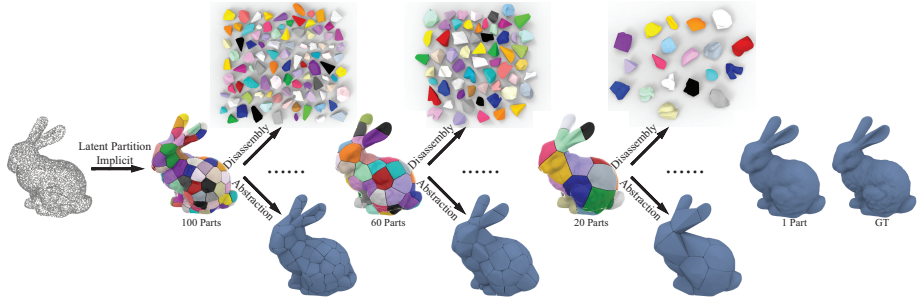}
  % replacing the above command with the one below will explicitly set
  % the bounding box of the PS figure to the rectangle (xl,yl),(xh,yh).
  % It will also prevent LaTeX from reading the PS file to determine
  % the bounding box (i.e., it will speed up the compilation process)
  % \includegraphics[width=.95\linewidth, bb=39 696 126 756]{sampleFig}
  %
  %
  \vspace{-0.32in}
\caption{\label{fig:Paris1}We introduce \textit{Latent Partition Implicit (LPI)} to represent 3D shapes. We learn LPI from 3D point clouds without requiring ground truth signed distances and point normals. We leverage a surface code to represent a part, and blend all parts with optional surface attributes, such as geodesic distance or segmentation, to reconstruct a surface, both of which are in the latent space. We can represent a shape (mesh) at different levels using different numbers of surface codes, which leads to different numbers of parts (meshes).}
\vspace{-0.3in}
\end{figure*}

Recent methods~\cite{DBLP:conf/cvpr/DengGYBHT20,Tretschk2020PatchNets,Genova:2019:LST,Genova_2020_CVPR} tried to resolve this problem in 3D spatial space. The key idea behind these methods is to learn to approximate a shape using various primitives, such as convex polytopes~\cite{DBLP:conf/cvpr/DengGYBHT20}, 3D Gaussian functions~\cite{Tretschk2020PatchNets,Genova:2019:LST,Genova_2020_CVPR}, superquadrics~\cite{Paschalidou2020CVPR} and homeomorphic mappings~\cite{DBLP:conf/cvpr/PaschalidouK0F21}. However, it is very hard to approximate shapes well by combining primitives together in the spatial space, even a large number of primitives can be used, since both primitive learning and primitive blending in the spatial space are still challenging.

To resolve this issue, we introduce a novel implicit representation to represent a single 3D shape as a set of parts in the latent space, towards both highly accurate and plausibly interpretable shape modeling. Our idea comes from the observation that we can not only learn part geometry but also blend parts together in the latent space, both of which are much easier to be implemented in the latent space than in the spatial space. For a 3D shape, we learn its Signed Distance Functions (SDFs) via blending a set of learnable latent codes in a geometry-aware way. Each latent code represents a part with a center on the surface, which we call \textit{surface codes}. Our insight of surface codes is that they enable us to flexibly leverage intrinsic attributes of shapes or additional properties of surfaces. Eventually, our learned SDFs can reconstruct not only the shape but also each part on the shape, both of which are plausible meshes. Our method partitions the global shape unity by casting the global shape modeling into multiple local part modeling, as demonstrated in Fig.~\ref{fig:Paris1}, which achieves important properties such as the reconstruction accuracy and convergence order inherited from the local part modeling. Therefore, we name our method \textit{Latent Partition Implicit (LPI)}. LPI is a multi-level representation for both rigid and non-rigid shapes, which can partition a shape into different numbers of parts after training. LPI can be learned without ground truth signed distances, point normals or any supervision for part partition. Our numerical and visual evaluation shows that LPI outperforms the state-of-the-art methods in terms of reconstruction accuracy and modeling interpretability. Our contributions are listed as follows.

\begin{enumerate}[i)]
\item We introduce LPI as a novel implicit representation for 3D shapes. It enables shape modeling and decomposition at the same time, which leads to highly accurate and plausibly interpretable shape modeling.
\item We justify the feasibility of learning part geometry and blending parts together in the latent space, which achieves more semantic and efficient parts representation.
\item Our method significantly outperforms the state-of-the-art methods in terms of reconstruction accuracy and modeling interpretability.
\end{enumerate}

\section{Related Work}
Deep learning-based 3D shape understanding~\cite{Zhu2021NICESLAM,ruckert2021adop,Park_2019_CVPR,MeschederNetworks,mildenhall2020nerf,Zhizhong2018seq,Zhizhong2019seq,3D2SeqViews19,wenxin_2020_CVPR,seqxy2seqzeccv2020paper,Zhizhong2018VIP,Zhizhong2020icml,Han2019ShapeCaptionerGCacmmm} has achieved very promising results in different tasks~\cite{zhizhongiccv2021finepoints,wenxincvpr2022,MAPVAE19,p2seq18,hutaoaaai2020,wenxin_2021a_CVPR,wenxin_2021b_CVPR,Jiang2019SDFDiffDRcvpr,zhizhongiccv2021completing,jain2021dreamfields,text2mesh,Novotny_2022_CVPR}.
%~\cite{Zhu2021NICESLAM,ruckert2021adop,Park_2019_CVPR,MeschederNetworks,mildenhall2020nerf,Zhizhong2018seq,Zhizhong2019seq,3D2SeqViews19,wenxin_2020_CVPR,seqxy2seqzeccv2020paper,Zhizhong2018VIP,Zhizhong2020icml,Han2019ShapeCaptionerGCacmmm,zhizhongiccv2021finepoints,wenxincvpr2022,MAPVAE19,p2seq18,hutaoaaai2020,wenxin_2021a_CVPR,wenxin_2021b_CVPR,Jiang2019SDFDiffDRcvpr,zhizhongiccv2021completing,tianyangcvpr2022,jain2021dreamfields,text2mesh,yu_and_fridovichkeil2021plenoxels,mueller2022instant,pococvpr2022,Peng2021SAP,sitzmann2019siren,Darmon_2022_CVPR,Hasselgren2022,Azinovic_2022_CVPR,Yu2022MonoSDF,predictivecontextpriors2022,onsurfacepriors2022}.
\noindent\textbf{Learning Global Implicit Functions. }Implicit functions have achieved remarkable results in geometry modeling. SDFs or occupancy fields can be learned using ground truth signed distances~\cite{DBLP:journals/corr/abs-1901-06802,Park_2019_CVPR,tianyangcvpr2022,pococvpr2022} and binary occupancy labels~\cite{MeschederNetworks,chen2018implicit_decoder}. The learned implicit functions leverage conditions to distinguish shape or scene identities. In different applications, conditions are latent codes obtained from different modalities, such as images for single image reconstruction~\cite{xu2019disn,pifuSHNMKL19,DBLP:conf/cvpr/ChibaneAP20,Gidi_2019_ICCV,Genova:2019:LST,seqxy2seqzeccv2020paper}, learnable latent codes for shape fitting~\cite{Park_2019_CVPR}, or point clouds for surface reconstruction~\cite{Williams_2019_CVPR,liu2020meshing,Mi_2020_CVPR,Genova:2019:LST,nipspoisson21,ErlerEtAl:Points2Surf:ECCV:2020,jiang2020lig,Liu2021MLS,Peng2021SAP,predictivecontextpriors2022,onsurfacepriors2022}.

With differentiable renderers~\cite{sitzmann2019srns,DIST2019SDFRcvpr,Jiang2019SDFDiffDRcvpr,prior2019SDFRcvpr,shichenNIPS,DBLP:journals/cgf/WuS20,Volumetric2019SDFRcvpr,lin2020sdfsrn,DBLP:journals/corr/abs-2104-10078,Darmon_2022_CVPR,Hasselgren2022,Azinovic_2022_CVPR,Yu2022MonoSDF,yu_and_fridovichkeil2021plenoxels,mueller2022instant,sitzmann2019siren}, we can also learn implicit functions with 2D supervision. This is achieved by minimizing the difference between the images rendered from the learned implicit functions and the ground truth images. Implicit functions achieve great results in geometry and color modeling for complex scenes with neural rendering~\cite{mildenhall2020nerf}.

Recent methods tried to learn implicit functions from 3D point clouds without ground truth signed distances or binary occupancy labels. Their contributions lied in additional constraints~\cite{gropp2020implicit,Atzmon_2020_CVPR,zhao2020signagnostic,atzmon2020sald,DBLP:journals/corr/abs-2106-10811,yifan2020isopoints} or the way of leveraging gradients to perceive the surroundings~\cite{Zhizhong2021icml,chibane2020neural} to learn signed~\cite{Zhizhong2021icml,gropp2020implicit,Atzmon_2020_CVPR,zhao2020signagnostic,atzmon2020sald,tang2021sign} or unsigned distance fields~\cite{chibane2020neural}.

\noindent\textbf{Learning Local Implicit Functions. }We can also learn implicit functions in local regions to capture more detailed geometry. The widely used strategy is to split the space occupied by the shape into a voxel grid~\cite{jiang2020lig,DBLP:conf/eccv/ChabraLISSLN20,Peng2020ECCV,DBLP:journals/corr/abs-2105-02788,takikawa2021nglod,Liu2021MLS}. Learnable latent codes located in voxels~\cite{jiang2020lig,DBLP:conf/eccv/ChabraLISSLN20,Peng2020ECCV,DBLP:journals/corr/abs-2105-02788} or vertices of voxels~\cite{takikawa2021nglod,Liu2021MLS}. The feature of each query is obtained using bilinear interpolation from these latent codes. Some other methods also used 3D Gaussian functions~\cite{Tretschk2020PatchNets,Genova_2020_CVPR} to cover the local regions.

\begin{figure*}[tb]
  \centering
  % the following command controls the width of the embedded PS file
  % (relative to the width of the current column)
  %\includegraphics[width=.95\linewidth, bb=39 696 126 756]{figures/definition3.eps}
   \includegraphics[width=\linewidth]{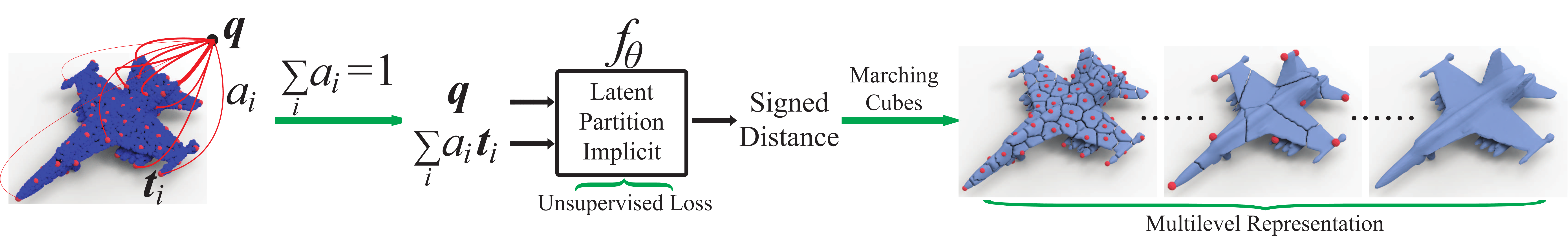}
  % replacing the above command with the one below will explicitly set
  % the bounding box of the PS figure to the rectangle (xl,yl),(xh,yh).
  % It will also prevent LaTeX from reading the PS file to determine
  % the bounding box (i.e., it will speed up the compilation process)
  % \includegraphics[width=.95\linewidth, bb=39 696 126 756]{sampleFig}
  %
  %
  \vspace{-0.32in}
\caption{\label{fig:overview}Demonstration of LPI. We can learn LPI from single point clouds without ground truth signed distances or point normals. LPI can represent shapes using different numbers of parts after training.}
\vspace{-0.25in}
\end{figure*}

The aforementioned methods can approximate the global implicit functions with~\cite{Tretschk2020PatchNets,Genova_2020_CVPR,giebenhain2021airnets,tianyangcvpr2022,pococvpr2022,Yao9607613} or without blending local implicit functions~\cite{jiang2020lig,DBLP:conf/eccv/ChabraLISSLN20,feng2022np}. Our method shares the idea of approximating global implicit functions by blending local ones, but our novelty lies in that we allow the neural network to adaptively split shapes so that we can blend parts with spatial surface properties or intrinsic attributes in the latent space well.

\noindent\textbf{Shape Decomposition. }Decomposing shapes into parts with specific attributes have been extensively studied in computer graphics~\cite{DBLP:journals/tog/MuntoniLSSP18,DBLP:journals/cgf/FilosciaAGMCC20,DBLP:journals/corr/abs-2112-13942,Yavartanoo_2021_ICCV,yao2021discovering}. Recent deep learning based methods tried to resolve this problem by learning primitives using a data-driven strategy~\cite{DBLP:conf/cvpr/DengGYBHT20,DBLP:conf/cvpr/PaschalidouGG20,DBLP:conf/cvpr/PaschalidouK0F21,DBLP:journals/corr/abs-2007-12944,DBLP:journals/corr/abs-2112-13942,3DIAS,9150964VoronoiNet}. The primitives could be convex polytopes~\cite{DBLP:conf/cvpr/PaschalidouGG20,DBLP:conf/cvpr/DengGYBHT20}, 3D Gaussian functions or spheres~\cite{Genova:2019:LST,DBLP:journals/corr/abs-2106-03804}. Instead of primitives learned by these methods, we can learn parts, and blend parts in the latent space rather than spatial space. This results in more accurate approximation and better interpretability for geometry modeling.

\section{Method}
\noindent\textbf{Problem Statement. }We aim to learn LPI as SDFs $f_{\theta}$ for a 3D shape $M$ using a deep neural network parameterized by $\theta$. We can use $f_{\theta}$ to describe multiple shapes with conditions represented as images or point clouds, but we only use single shapes without conditions to simplify the technical details. Besides this, we also aim to decompose shape $M$ into $I$ parts $p_i$, each of which is also represented as an SDF, where $i\in[1,I]$. To achieve this, we aim to learn $f_{\theta}$ using the following equation,\vspace{-0.15in}

\begin{equation}
\label{eq:1}
\begin{aligned}
f_{\theta}(\bm{q},\bm{a})=s,
\end{aligned}
\end{equation}

\noindent where $f_{\theta}$ provides a signed distance $s$ for a query $\bm{q}$ according to an affinity vector $\bm{a}$. Each element $a_i$ in $\bm{a}$ represents its similarity to each part $p_i$ in terms of some metrics, and $\sum_{i\in[1,I]}a_i=1$. The affinity vector $\bm{a}$ determines how we can blend local SDFs in the latent space into a global SDF at the query location $\bm{q}$.

\noindent\textbf{Overview. }We demonstrate LPI in Fig.~\ref{fig:overview}. Given a 3D point cloud representing shape $M$, we first sample $I$ region centers $\bm{r}_i$ on the point cloud using farthest point sampling (FPS). Each center $\bm{r}_i$ localizes a part $p_i$. We represent each part $p_i$ using a latent code $\bm{t}_i\in\mathbb{R}^{1\times T}$ in the latent space. The latent codes $\bm{t}_i$ associated with part centers on the surface are called surface codes. $\bm{t}_i$ can be used to produce a local SDF to describe the surface of part $p_i$. We blend all parts together in the latent space by weighting latent codes $\bm{t}_i$ using an affinity vector $\bm{a}$ at each query location $\bm{q}$, which results in a global SDF. We will introduce how to obtain the affinity vector for different kinds of shapes in different applications later. Eventually, we learn the SDFs $f_{\theta}$ which can be further used to reconstruct the surface and decompose the shape into parts at multiple levels.

\begin{wrapfigure}{r}{0.7\linewidth}% 驴驴脦脛脳脰脛脷脠脻碌脛脳贸虏脿
\vspace{-0.3in}
\includegraphics[width=\linewidth]{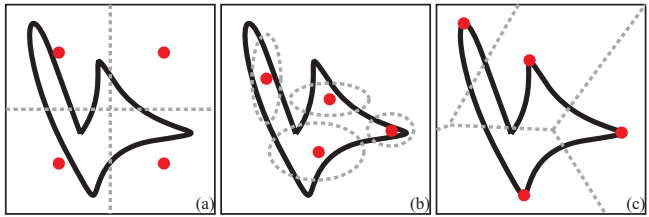}
\vspace{-0.3in}
\caption{\label{fig:Surfacepoint} Regions covered by voxels and ellipsoids in (a) and (b). Our regions are covered by Voronoi cells in (c). Red points are region centers.}
\vspace{-0.28in}
\end{wrapfigure}

\noindent\textbf{Surface Codes. }It is not new to cast the learning of global SDFs into the learning of multiple local SDFs. The rationale behind this is that local regions are simpler than global shapes, which is easier to learn. A widely used strategy is to regard the space holding 3D shapes as a voxel grid~\cite{jiang2020lig,DBLP:conf/eccv/ChabraLISSLN20,Peng2020ECCV} or a 3D Gaussian mixture model~\cite{Genova:2019:LST,DBLP:journals/corr/abs-2106-03804}, and use a latent code to cover each split region, such as voxels or ellipsoids shown in Fig.~\ref{fig:Surfacepoint} (a) and (b). Different from these methods, we uniformly distribute region centers on the input point cloud, and use latent codes $\bm{t}_i$ to cover each split region as a Voronoi cell in Fig.~\ref{fig:Surfacepoint} (c). The benefits we can get are three-folds, which leads our method to be geometry-aware. One benefit is that these centers enable us to perceive the intrinsic attributes of surface, such as geodesic distances, especially for non-rigid shapes, such as humans. This would be helpful to plausibly decompose non-rigid shapes. Another benefit is that Voronoi cells could decompose shapes in a more compact way than voxel grids, especially with the same number of latent codes, which results in more uniformly decomposed parts. This is very helpful to achieve highly accurate shape abstraction without losing too much structure information when abstracting decomposed parts into convex polytopes. Moreover, we can also flexibly leverage additional properties of the surface, such as segmentation, to produce more semantic parts in surface reconstruction. Surface codes $\bm{t}_i$ are learnable parameters for each shape. They are learned with other network parameters $\theta$.

%\begin{figure}[tb]
%  \centering
%  % the following command controls the width of the embedded PS file
%  % (relative to the width of the current column)
%  %\includegraphics[width=.95\linewidth, bb=39 696 126 756]{figures/definition3.eps}
%   \includegraphics[width=\linewidth]{SplitEPS-eps-converted-to.pdf}
%  % replacing the above command with the one below will explicitly set
%  % the bounding box of the PS figure to the rectangle (xl,yl),(xh,yh).
%  % It will also prevent LaTeX from reading the PS file to determine
%  % the bounding box (i.e., it will speed up the compilation process)
%  % \includegraphics[width=.95\linewidth, bb=39 696 126 756]{sampleFig}
%  %
%  %
%  \vspace{-0.3in}
%\caption{\label{fig:Surfacepoint} Regions covered by voxels and ellipsoids in (a) and (b). Our regions are covered by Voronoi cells in (c). Red points are region centers.}
%\vspace{-0.28in}
%\end{figure}

\noindent\textbf{Blending Regions in the Latent Space. }However, the outputs of neighboring local functions are inconsistent. This problem significantly slows down the convergence during training, and also affects the final reconstruction accuracy. Current methods resolve this problem by blending regions into a global shape in the spatial space~\cite{OhtakeBATS03,Tretschk2020PatchNets}. They mainly leverage a weight function with a local support to blend the output of multiple local functions, such as local SDFs. This will also bring another issue, that is, the blending requires all neighboring local functions to produce the prediction, which is not affordable when there are lots of local functions. Therefore, DeepLS~\cite{DBLP:conf/eccv/ChabraLISSLN20} does not blend local functions but increases the receptive field of each local region to keep region borders consistent.

%\begin{figure}[tb]
%  \centering
%  % the following command controls the width of the embedded PS file
%  % (relative to the width of the current column)
%  %\includegraphics[width=.95\linewidth, bb=39 696 126 756]{figures/definition3.eps}
%   \includegraphics[width=\linewidth]{Euclidean-eps-converted-to.pdf}
%  % replacing the above command with the one below will explicitly set
%  % the bounding box of the PS figure to the rectangle (xl,yl),(xh,yh).
%  % It will also prevent LaTeX from reading the PS file to determine
%  % the bounding box (i.e., it will speed up the compilation process)
%  % \includegraphics[width=.95\linewidth, bb=39 696 126 756]{sampleFig}
%  %
%  %
%  \vspace{-0.3in}
%\caption{\label{fig:Euclidean} (a) Euclidean distance $d_E$ between query $\bm{q}$ and region centers $\bm{r}_i$. (b) Intrinsic distance $d_G$ between query $\bm{q}$ and region centers $\bm{r}_i$.}
%\vspace{-0.28in}
%\end{figure}

We resolve this problem by blending local functions in the latent space. Rather than blending outputs in the spatial space, we blend latent codes $\bm{t}_i$ corresponding to all local SDFs at each query $\bm{q}$ into a weighted latent code $\bm{w}$. $\bm{w}$ is further concatenated with query $\bm{q}$ to produce its signed distance. We leverage an affinity vector $\bm{a}$ at each query $\bm{q}$ to weight latent codes $\bm{t}_i$ linearly into $\bm{w}$ below,\vspace{-0.1in}

\begin{equation}
\label{eq:2}
\begin{aligned}
\bm{w}=\sum_{i\in[1,I]}a_i\bm{t}_i,
\vspace{-0.1in}
\end{aligned}
\end{equation}

\noindent where $a_i$ represents the affinity between query $\bm{q}$ and part $p_i$ in terms of some metrics, and we are very flexible to define the affinity metrics, such as Euclidean distance, intrinsic distance, or semantic distance, which achieves different shape decompositions. This flexibility brought by surface codes also differentiates our method with other ones that use bilinear interpolation to blend latent codes on voxel grids~\cite{takikawa2021nglod,DBLP:journals/corr/abs-2105-02788}, especially for non-rigid shapes like humans.

\begin{wrapfigure}{r}{0.7\linewidth}% 驴驴脦脛脳脰脛脷脠脻碌脛脳贸虏脿
\vspace{-0.3in}
\includegraphics[width=\linewidth]{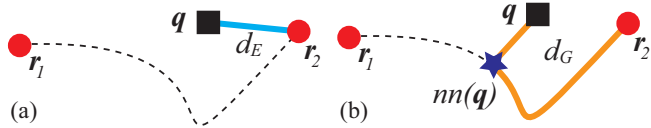}
\vspace{-0.3in}
\caption{\label{fig:Euclidean} (a) Euclidean distance $d_E$ between query $\bm{q}$ and region centers $\bm{r}_i$. (b) Intrinsic distance $d_G$ between query $\bm{q}$ and region centers $\bm{r}_i$.}
\vspace{-0.22in}
\end{wrapfigure}

\noindent\textbf{Affinity Vector $\bm{a}$. }One benefit that we can have with surface codes is to obtain geometry awareness. We can encode the geometry by flexibly leveraging different similarity metrics to form the affinity vector $\bm{a}$ for each query $\bm{q}$. We will introduce three metrics including Euclidean distance $d_E$, intrinsic distance $d_G$, and semantic distance $d_S$ to evaluate the distance between query $\bm{q}$ and each part $p_i$.

\begin{wrapfigure}{r}{0.5\linewidth}% 驴驴脦脛脳脰脛脷脠脻碌脛脳贸虏脿
\vspace{-0.4in}
\includegraphics[width=\linewidth]{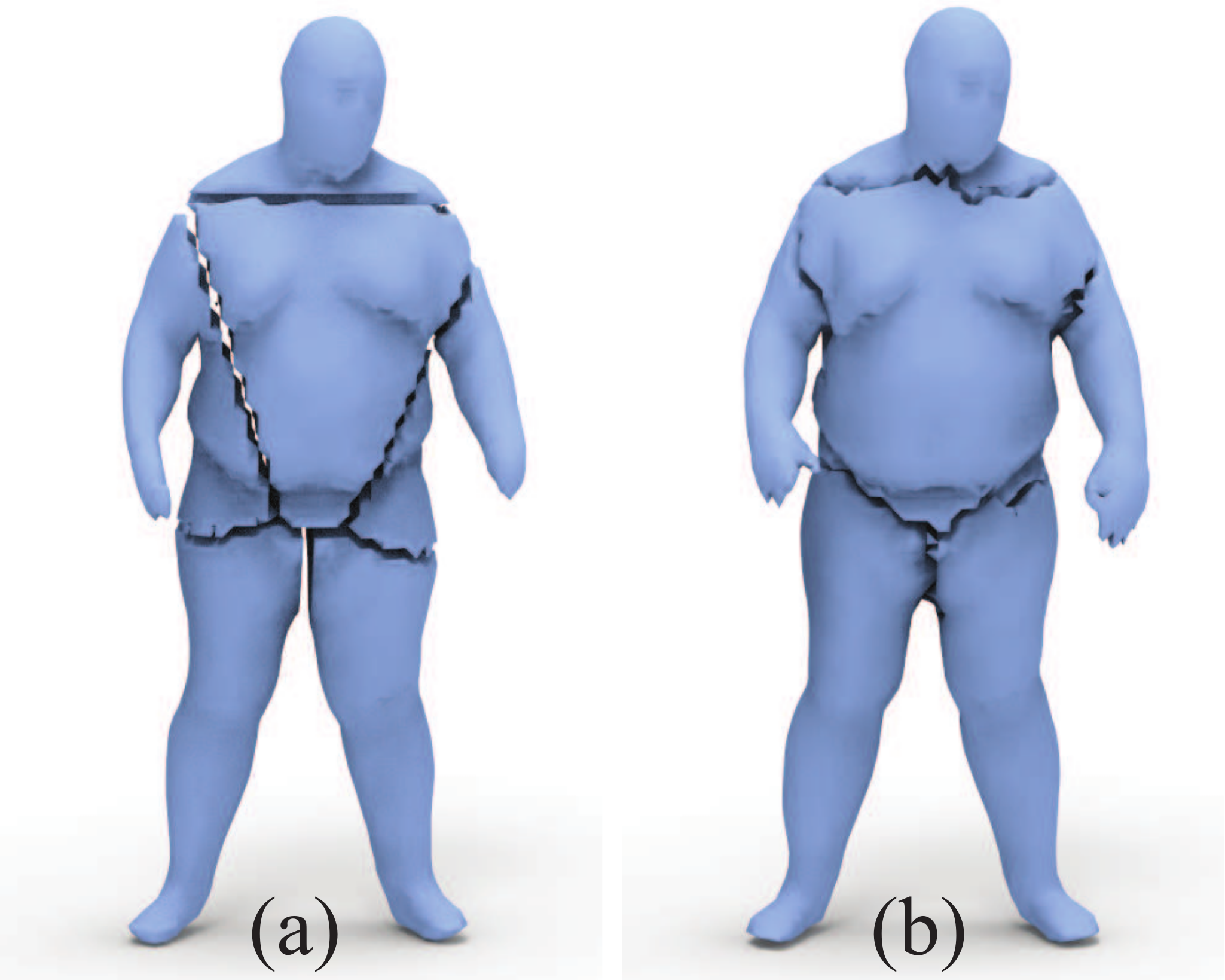}
\vspace{-0.4in}
\caption{\label{fig:DistanceComp}The effect of Euclidean distance $d_E$ and intrinsic distance $d_G$}
\vspace{-0.18in}
\end{wrapfigure}

\noindent\textbf{1. Euclidean distance $d_E$. }The most intuitive metric is Euclidean distance, the distance between query $\bm{q}$ and part $p_i$ can be evaluated using the equation below,\vspace{-0.15in}

\begin{equation}
\label{eq:3}
\begin{aligned}
d_E(\bm{q},p_i)=||\bm{q}-\bm{r}_i||_2,
\end{aligned}
\end{equation}

\noindent where $\bm{r}_i$ is the center of region $p_i$. This metric indicates that query $\bm{q}$ is more related to a region $p_i$ if $\bm{q}$ is nearer to its center $\bm{r}_i$, as demonstrated in Fig.~\ref{fig:Euclidean} (a). However, Euclidean distance does not care about the intrinsic property of shapes, especially on non-rigid shapes.

\noindent\textbf{2. Intrinsic distance $d_G$. }To resolve this issue, we also introduce intrinsic distance $d_G$ as a distance metric. The intrinsic distance is formed by Euclidean distance and geodesic distance. The geodesic distance is the distance of the shortest path connecting two points on the surface. So, the intrinsic distance between query $\bm{q}$ and part $p_i$ can be evaluated using the equation below,\vspace{-0.15in}

\begin{equation}
\label{eq:4}
\begin{aligned}
d_G(\bm{q},p_i)=||\bm{q}-nn(\bm{q},M)||_2+G(nn(\bm{q},M)),\bm{r}_i),
\end{aligned}
\end{equation}

\noindent where $G$ represents all pair-wise geodesic distances on surface $M$, $nn(\bm{q},M)=\argmin_{\bm{q}'\in M}||\bm{q}-\bm{q}'||_2$ is the nearest point on $M$ of query $\bm{q}$. We demonstrate $d_G$ in Fig.~\ref{fig:Euclidean} (b). We first project query $\bm{q}$ to surface $M$ by finding the nearest point $nn(\bm{q},M)$ of $\bm{q}$, then we calculate the geodesic distance between $nn(\bm{q},M)$ and region center $\bm{r}_i$. In experiments, we calculate the geodesic distance between each two points on $M$ using heat method~\cite{Crane:2017:HMD} in advance.

We highlight the difference between Euclidean distance $d_E$ and intrinsic distance $d_G$ on a non-rigid shape in Fig.~\ref{fig:DistanceComp}. Using the same set of six region centers $\bm{r}_i$, LPI with $d_E$ splits the human body without considering the intrinsic structures in Fig.~\ref{fig:DistanceComp} (a), while LPI with $d_G$ can produce more reasonable splitting using geodesic distances.

For both $d_E$ and $d_G$, we use the Gaussian function to obtain the similarity $a_i$ between query $\bm{q}$ and part $p_i$, and the similarity is normalized by the sum of all similarities to all parts,\vspace{-0.13in}

\begin{equation}
\label{eq:51}
\begin{aligned}
a_i=e^{-d(\bm{q},p_i)/\sigma}/\sum_{i'\in[1,I]}e^{-d(\bm{q},p_{i'})/\sigma},
\end{aligned}
\end{equation}

\noindent where $\sigma$ is a decay parameter that determines the range in which surface codes get involved, and $d$ could be either $d_E$ or $d_G$.

%\begin{wrapfigure}{r}{0.7\linewidth}% 驴驴脦脛脳脰脛脷脠脻碌脛脳贸虏脿
%\includegraphics[width=\linewidth]{Segmentation-eps-converted-to.pdf}
%\vspace{-0.3in}
%\caption{\label{fig:segmentation}LPI can reconstruct surfaces and semantic parts in (b) according to segmentation.}
%\vspace{-0.22in}
%\end{wrapfigure}

\begin{wrapfigure}{r}{0.7\linewidth}% 驴驴脦脛脳脰脛脷脠脻碌脛脳贸虏脿
\vspace{-0.35in}
\includegraphics[width=\linewidth]{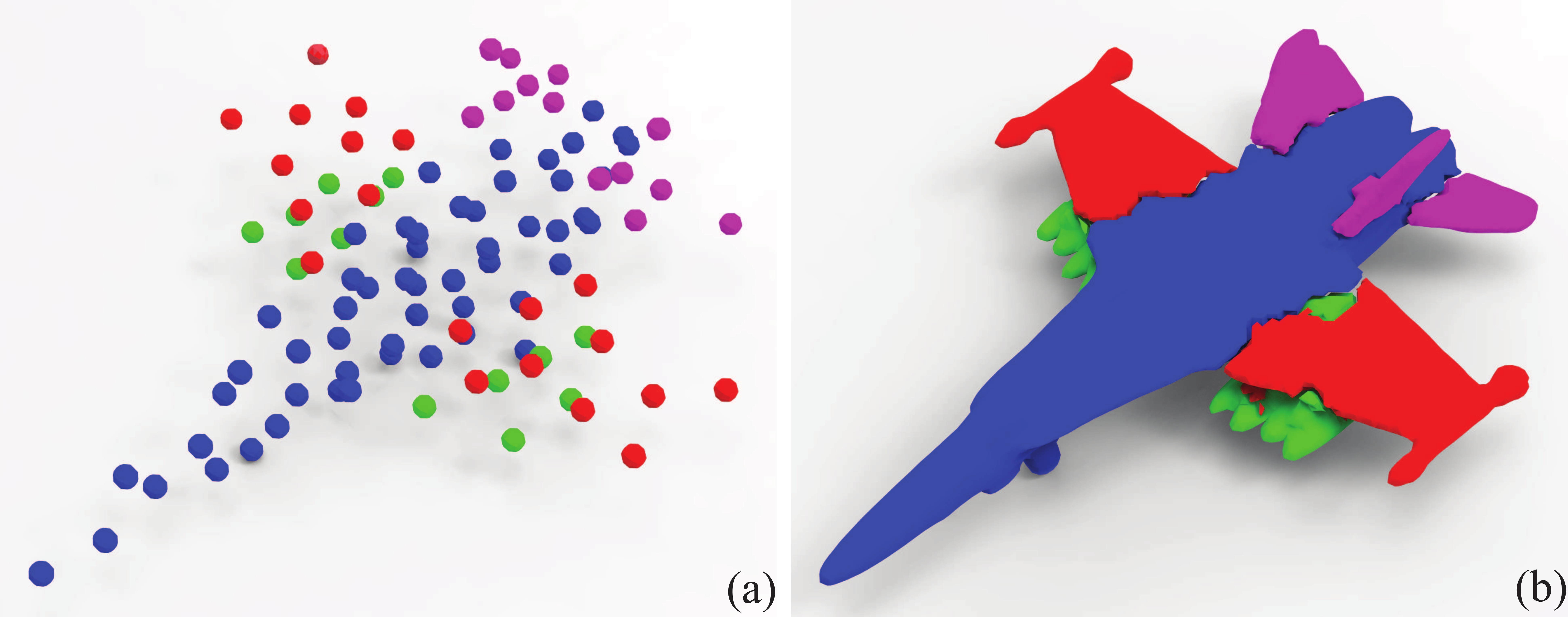}
\vspace{-0.3in}
\caption{\label{fig:segmentation}LPI can reconstruct surfaces and semantic parts in (b) according to segmentation.}
\vspace{-0.35in}
\end{wrapfigure}

\noindent\textbf{3. Semantic distance $d_S$. }With additional properties of the surface, such as segmentation information, we can also obtain the affinity vector using semantic distance $d_S$. In this case, $d_S$ is an indicator which indicates the segment that query $\bm{q}$ belongs to. In this way, the affinity vector indicates the semantic similarity between a query and each predefined segment rather than region centers. Assume we were given a sparse point cloud $E$ containing $4$ segments in Fig.~\ref{fig:segmentation} (a), we find the nearest point $nn(\bm{q},E)$ on $E$ for query $\bm{q}$, and label $\bm{q}$ using the same segment label of $nn(\bm{q},E)$. Then, we set the affinity between $\bm{q}$ and its segment label to $0.8$ while set the affinities to other $3$ segments uniformly to keep $\sum_{i\in[1,T]}a_i=1$. With the segmentation as guidance, LPI can reconstruct the surface of the point cloud in Fig.~\ref{fig:overview} (a) using $4$ semantic parts shown in Fig.~\ref{fig:segmentation} (b).

%\begin{figure}[tb]
%  \centering
%  % the following command controls the width of the embedded PS file
%  % (relative to the width of the current column)
%  %\includegraphics[width=.95\linewidth, bb=39 696 126 756]{figures/definition3.eps}
%   \includegraphics[width=\linewidth]{Segmentation-eps-converted-to.pdf}
%  % replacing the above command with the one below will explicitly set
%  % the bounding box of the PS figure to the rectangle (xl,yl),(xh,yh).
%  % It will also prevent LaTeX from reading the PS file to determine
%  % the bounding box (i.e., it will speed up the compilation process)
%  % \includegraphics[width=.95\linewidth, bb=39 696 126 756]{sampleFig}
%  %
%  %
%\caption{\label{fig:segmentation} }
%\end{figure}

\noindent\textbf{Loss Function. }We learn LPI as SDFs $f_{\theta}(\bm{q},\bm{a})$ without ground truth signed distances or point normals by optimizing network parameters $\theta$ and all surface codes $\{\bm{t}_i\}$. We leverage a loss function that is modified from the pulling loss introduced in~\cite{Zhizhong2021icml} to resist the sparseness of input point clouds. The loss function is defined below,\vspace{-0.1in}

\begin{equation}
\label{eq:5}
\begin{aligned}
\min_{\theta,\{\bm{t}_i\}} (\sum_{\bm{x}\in M}\min_{\bm{y}\in \{\bm{q}'\}} ||\bm{x}-\bm{y}||_2^2+\sum_{\bm{y}\in \{\bm{q}'\}}\min_{\bm{x}\in M}||\bm{x}-\bm{y}||_2^2),
\end{aligned}
\end{equation}

\noindent where we obtain the points $\{\bm{q}'\}$ by projecting each query $\bm{q}$ using the predicted signed distance $s=f_{\theta}(\bm{q},\bm{a})$ and the gradient $\nabla f_{\theta}(\bm{q},\bm{a})$, such that $\bm{q}'=\bm{q}-s\nabla f_{\theta}(\bm{q},\bm{a})/||\nabla f_{\theta}(\bm{q},\bm{a})||_2$.

%If the ground truth signed distances are available, we can train our network by minimizing a mean square error (MSE) as a regression problem.

\begin{wrapfigure}{r}{0.7\linewidth}% 驴驴脦脛脳脰脛脷脠脻碌脛脳贸虏脿
\vspace{-0.35in}
\includegraphics[width=\linewidth]{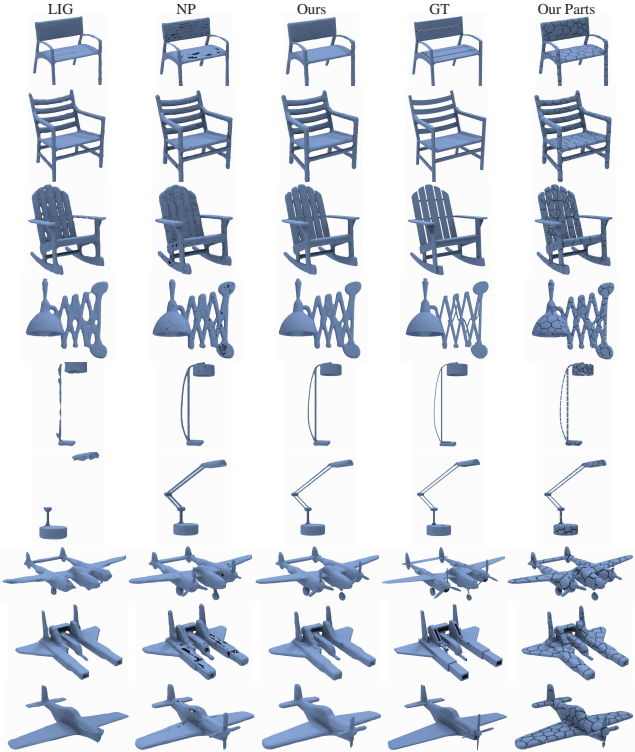}
\vspace{-0.3in}
\caption{\label{fig:shapenet}Visual comparison with LIG~\cite{jiang2020lig} and NP~\cite{Zhizhong2021icml} in surface reconstruction under ShapeNet.}
\vspace{-0.35in}
\end{wrapfigure}

\noindent\textbf{Shape and Part Reconstruction. }With the learned SDFs $f_{\theta}$, we reconstruct the shape surface by evaluating
$f_{\theta}$ at points on a regular grid and running the marching cubes~\cite{Lorensen87marchingcubes}. To reconstruct each part centered at $\bm{r}_i$, we evaluate $f_{\theta}$ at points on the regular grid whose nearest region center is $\bm{r}_i$ while leveraging an unseen weighted latent code $\bm{w}$ to predict signed distances at the rest points on the regular grid, and we also leverage marching cubes to reconstruct the surface of each part. After training, we can down sample region centers $\{\bm{r}_i\}$ to reconstruct the shape with fewer parts in the same way.\vspace{-0.2in}

\section{Experiments}\vspace{-0.15in}
\subsection{Setup}\vspace{-0.1in}
\noindent\textbf{Details. }For fair comparisons, we leverage the same neural network as NeuralPull (NP)~\cite{Zhizhong2021icml} to learn the SDFs $f_{\theta}$. Moreover, we overfit single shapes in an unsupervised scenario using Eq.~(\ref{eq:5}) without requiring signed distances or point normals, or in a supervised scenario using an MSE loss with signed distance supervision for fair comparisons with others. In an unsupervised scenario, we leverage point clouds as input and sample queries around the point clouds using the same method as NP~\cite{Zhizhong2021icml}. In a supervised scenario, we establish a training set containing queries and their ground truth signed distances for each shape. We sample queries and calculate the ground truth signed distances around the shape using the same method as DSDF~\cite{Park_2019_CVPR}. All compared methods use the same training set. In both scenarios, we leverage the marching cubes algorithm~\cite{Lorensen87marchingcubes} to reconstruct surfaces using the learned SDFs. We set $\sigma=1$ in Eq.~(\ref{eq:51}) and the dimension of $\bm{t}_i$ is $T=100$.

%\begin{figure}[tb]
%  \centering
%  % the following command controls the width of the embedded PS file
%  % (relative to the width of the current column)
%  %\includegraphics[width=.95\linewidth, bb=39 696 126 756]{figures/definition3.eps}
%   \includegraphics[width=0.5\linewidth]{ShapeNetEPS-eps-converted-to.pdf}
%  % replacing the above command with the one below will explicitly set
%  % the bounding box of the PS figure to the rectangle (xl,yl),(xh,yh).
%  % It will also prevent LaTeX from reading the PS file to determine
%  % the bounding box (i.e., it will speed up the compilation process)
%  % \includegraphics[width=.95\linewidth, bb=39 696 126 756]{sampleFig}
%  %
%  %
%  \vspace{-0.3in}
%\caption{\label{fig:shapenet}Visual comparison with LIG~\cite{jiang2020lig} and NP~\cite{Zhizhong2021icml} in surface reconstruction under ShapeNet.}
%\vspace{-0.3in}
%\end{figure}

\noindent\textbf{Datasets. }We evaluate our method in three datasets including ShapeNet~\cite{shapenet2015}, FAMOUS~\cite{ErlerEtAl:Points2Surf:ECCV:2020} and D-FAUST~\cite{dfaust:CVPR:2017}. We leverage the same subset of ShapeNet with the same train/test splitting as~\cite{Zhizhong2021icml,liu2020meshing}. The subset of ShapeNet contains 8 shape classes. FAMOUS is a dataset released by Points2Surf~\cite{ErlerEtAl:Points2Surf:ECCV:2020}, it contains 22 well-known 3D shapes. D-FAUST is a large-scale dataset containing human meshes. We randomly select 10 shapes and 100 shapes representing different human identities and poses as two sets.

%We leverage ShapeNet and FAMOUS to evaluate our ability of representing rigid shapes while leveraging D-FAUST to evaluate our ability of representing non-rigid shapes.

\noindent\textbf{Metrics. }We follow MeshingPoint~\cite{liu2020meshing} to leverage L2 Chamfer Distance (L2CD), Normal Consistency (NC)~\cite{MeschederNetworks} and F-score~\cite{Tatarchenko_2019_CVPR} to evaluate our surface reconstruction accuracy under ShapeNet. We sample $100K$ points on the reconstructed and the ground truth surfaces to calculate the L2CD. Following Points2Surf~\cite{ErlerEtAl:Points2Surf:ECCV:2020}, we leverage L2CD to evaluate the reconstruction error between our reconstructed surfaces and the ground truth meshes under FAMOUS dataset. We sample $10K$ points on both reconstructed surfaces and the ground truth meshes to calculate the L2CD. Under D-FAUST dataset, we leverage L1 Chamfer Distance (L1CD), L2CD, NC and IoU to evaluate the reconstruction accuracy. We also sample $10K$ points to calculate CD.\vspace{-0.2in}

\subsection{Surface Reconstruction}\vspace{-0.1in}
\noindent\textbf{ShapeNet. }We first evaluate our method in surface reconstruction under ShapeNet. We train our method to reconstruct the surface from each point cloud without using the signed distance supervision. We leverage Euclidean distance $d_E$ to evaluate the distance between a query and each one of $I=100$ regions to obtain the affinity vector $\bm{a}$. We use FPS to randomly sample the $I=100$ region centers on the input point clouds.

\begin{table}[tb]
\centering
  % ????????Comparison of shape reconstruction with known camera pose from silhouette images with different resolutions in terms of CD
\resizebox{\linewidth}{!}{
    \begin{tabular}{c|c|c|c|c|c|c|c|c|c|c|c|c}  % ?????
     \hline
       %\cline{1-12}
       %\hline
        Class& PSR& DMC & BPA & ATLAS &DMC&DSDF& DGP &MeshP&NUD&SALD&NP&Ours \\  % ????? 脗搂脙隆?
     \hline
        Display& 0.273& 0.269 & 0.093 & 1.094 &0.662&0.317& 0.293 & 0.069 & 0.077 &-& 0.039&\textbf{0.0080}\\
        Lamp &0.227&0.244&0.060&1.988&3.377&0.955&0.167&0.053&0.075&0.071&0.080&\textbf{0.0172}\\
        Airplane&0.217&0.171&0.059&1.011&2.205&1.043&0.200&0.049&0.076&0.054&0.008&\textbf{0.0060}\\
        Cabinet&0.363&0.373&0.292&1.661&0.766&0.921&0.237&0.112&0.041&-&0.026&\textbf{0.0179}\\
        Vessel&0.254&0.228&0.078&0.997&2.487&1.254&0.199&0.061&0.079&-&0.022&\textbf{0.0092}\\
        Table&0.383&0.375&0.120&1.311&1.128&0.660&0.333&0.076&0.067&0.066&0.060&\textbf{0.0436}\\
        Chair&0.293&0.283&0.099&1.575&1.047&0.483&0.219&0.071&0.063&0.061&0.054&\textbf{0.0187}\\
        Sofa&0.276&0.266&0.124&1.307&0.763&0.496&0.174&0.080&0.071&0.058&0.012&\textbf{0.0164}\\
     \hline
     Mean&0.286&0.276&0.116&1.368&1.554&0.766&0.228&0.071&0.069&0.062&0.038&\textbf{0.0171}\\
     \hline
   \end{tabular}}
   %\vspace{-0.15in}
   \caption{L2CD ($\times100$) comparison under ShapeNet.}
   \vspace{-0.45in}
   \label{table:t10}
\end{table}

We compare our method with the classic surface reconstruction methods and the state-of-the-art data-driven based methods. The compared methods include PSR~\cite{journals/tog/KazhdanH13}, Ball-Pivoting algorithm (BPA)~\cite{817351TVCG}, ATLAS~\cite{Groueix_2018_CVPR}, Deep Geometric Prior (DGP)~\cite{Williams_2019_CVPR}, Deep Marching Cube (DMC)~\cite{Liao2018CVPR}, DeepSDF (DSDF)~\cite{Park_2019_CVPR}, MeshP~\cite{liu2020meshing}, Neural Unsigned Distance (NUD)~\cite{chibane2020neural}, SALD~\cite{atzmon2020sald}, Local SDF (GRID)~\cite{jiang2020lig}, IMNET~\cite{chen2018implicit_decoder} and NeuralPull (NP)~\cite{Zhizhong2021icml}.

\begin{table}[tb]
\centering
 % ????????Comparison of shape reconstruction with known camera pose from silhouette images with different resolutions in terms of CD
\resizebox{\linewidth}{!}{
    \begin{tabular}{c|c|c|c|c|c|c|c|c|c|c|c}  % ?????
     \hline
       %\cline{1-12}
       %\hline
        Class& PSR& DMC & BPA & ATLAS &DMC&DSDF&MeshP&LIG&IMNET&NP&Ours \\  % ?????茂驴陆茂驴陆?
     \hline
        Display& 0.889& 0.842 & 0.952 & 0.828 &0.882&0.932& 0.974 & 0.926 &0.574&0.964&\textbf{0.9780}\\
        Lamp &0.876&0.872&0.951&0.593&0.725&0.864&\textbf{0.963}&0.882&0.592&0.930&0.9503\\
        Airplane&0.848&0.835&0.926&0.737&0.716&0.872&0.955&0.817&0.550&0.947&\textbf{0.9560}\\
        Cabinet&0.880&0.827&0.836&0.682&0.845&0.872&0.957&0.948&0.700&0.930&\textbf{0.9576}\\
        Vessel&0.861&0.831&0.917&0.671&0.706&0.841&0.953&0.847&0.574&0.941&\textbf{0.9564}\\
        Table&0.833&0.809&0.919&0.783&0.831&0.901&\textbf{0.962}&0.936&0.702&0.908&0.9527\\
        Chair&0.850&0.818&0.938&0.638&0.794&0.886&\textbf{0.962}&0.920&0.820&0.937&0.9545\\
        Sofa&0.892&0.851&0.940&0.633&0.850&0.906&\textbf{0.971}&0.944&0.818&0.951&\textbf{0.9713}\\
     \hline
     Mean&0.866&0.836&0.923&0.695&0.794&0.884&\textbf{0.962}&0.903&0.666&0.939&0.9596\\
     \hline
   \end{tabular}}
   %\vspace{-0.15in}
   \caption{Normal consistency comparison under ShapeNet.}
   \vspace{-0.25in}
   \label{table:t11}
\end{table}

We report the numerical comparison in terms of L2CD in Tab.~\ref{table:t10}, NC in Tab.~\ref{table:t11}, F-score with $\mu$ in Tab.~\ref{table:st12}, F-score with $2\mu$ in Tab.~\ref{table:t13}. Our method achieves the best in terms of L2CD and F-score, while obtains the state-of-the-art in terms of NC. We further highlight our advantage in the visual comparison with LIG and NP in Fig.~\ref{fig:shapenet}. The comparison shows that our method can reconstruct surfaces with complex geometry in higher accuracy. We also visualize the parts that each surface code covers on the shape, which demonstrates that our method can reconstruct both plausible shapes and parts as meshes.

\begin{table}[tb]
\vspace{-0.05in}
\centering
  % ????????Comparison of shape reconstruction with known camera pose from silhouette images with different resolutions in terms of CD
\resizebox{\linewidth}{!}{
    \begin{tabular}{c|c|c|c|c|c|c|c|c|c|c|c|c|c}  % ?????
     \hline
       %\cline{1-12}
       %\hline
        Class& PSR& DMC & BPA & ATLAS &DMC&DSDF& DGP &MeshP&NUD&LIG&IMNET&NP&Ours \\  % ?????茂驴陆茂驴陆?
     \hline
        Display&0.468& 0.495& 0.834& 0.071& 0.108& 0.632& 0.417& 0.903& 0.903&0.551&0.601&0.989&\textbf{0.9978}\\
        Lamp& 0.455 &0.518& 0.826& 0.029 &0.047& 0.268& 0.405& 0.855&0.888&0.624&0.836&0.891&\textbf{0.9889}\\
        Airplane&0.415 &0.442& 0.788& 0.070& 0.050& 0.350& 0.249 &0.844&0.872&0.564&0.698&0.996&\textbf{0.9989}\\
        Cabinet&0.392 &0.392& 0.553& 0.077& 0.154 &0.573& 0.513& 0.860&0.950&0.733&0.343&0.980&\textbf{0.9849}\\
        Vessel&0.415& 0.466& 0.789& 0.058& 0.055& 0.323& 0.387& 0.862&0.883&0.467&0.147&0.985&\textbf{0.9955}\\
        Table&0.233& 0.287& 0.772& 0.080 &0.095& 0.577& 0.307& 0.880& 0.908&0.844&0.425&0.922&\textbf{0.9789}\\
        Chair&0.382 &0.433& 0.802& 0.050& 0.088& 0.447& 0.481& 0.875& 0.913&0.710&0.181&0.954&\textbf{0.9897}\\
        Sofa&0.499& 0.535& 0.786& 0.058& 0.129& 0.577& 0.638& 0.895&0.945&0.822&0.199&0.968&\textbf{0.9946}\\
     \hline
     Mean&0.407 &0.446 &0.769& 0.062 &0.091& 0.468& 0.425& 0.872&0.908&0.664&0.429&0.961&\textbf{0.9912}\\
     \hline
   \end{tabular}}
   %\vspace{-0.15in}
   \caption{F-score($\mu$) comparison under ShapeNet. $\mu=0.002$.}
   \vspace{-0.45in}
   \label{table:st12}
\end{table}

\begin{table}[tb]
\vspace{-0.3in}
\centering
  % ????????Comparison of shape reconstruction with known camera pose from silhouette images with different resolutions in terms of CD
\resizebox{\linewidth}{!}{
    \begin{tabular}{c|c|c|c|c|c|c|c|c|c|c|c}  % ?????
     \hline
       %\cline{1-12}
       %\hline
        Class& PSR& DMC & BPA & ATLAS &DMC&DSDF& DGP &MeshP&NUD&NP&Ours \\  % ?????茂驴陆茂驴陆?
     \hline
        Display&0.666& 0.669& 0.929& 0.179& 0.246& 0.787& 0.607& 0.975&0.944&0.991&\textbf{0.9993}\\
        Lamp&0.648& 0.681& 0.934& 0.077& 0.113& 0.478& 0.662& 0.951&0.945&0.924&\textbf{0.9954}\\
        Airplane&0.619& 0.639& 0.914& 0.179 &0.289& 0.566& 0.515& 0.946&0.944&0.997&\textbf{0.9998}\\
        Cabinet&0.598& 0.591& 0.706& 0.195& 0.128& 0.694& 0.738& 0.946&0.980&0.989&\textbf{0.9938}\\
        Vessel&0.633& 0.647& 0.906& 0.153& 0.120& 0.509& 0.648& 0.956&0.945&0.990&\textbf{0.9985}\\
        Table&0.442& 0.462& 0.886& 0.195& 0.221& 0.743& 0.494& 0.963&0.922&0.973&\textbf{0.9866}\\
        Chair&0.617& 0.615& 0.913& 0.134& 0.345& 0.665 & 0.693& 0.964&0.954&0.969&\textbf{0.9940}\\
        Sofa&0.725& 0.708& 0.895& 0.153& 0.208& 0.734& 0.834& 0.972&0.968&0.974&\textbf{0.9982}\\
     \hline
     Mean&0.618& 0.626& 0.885& 0.158& 0.209& 0.647& 0.649 &0.959&0.950&0.976&\textbf{0.9957}\\
     \hline
   \end{tabular}}
   %\vspace{-0.15in}
   \caption{F-score($2\mu$) comparison under ShapeNet. $\mu=0.002$.}
   \vspace{-0.47in}
   \label{table:t13}
\end{table}

\begin{wraptable}{r}{0.5\linewidth}% 靠文字内容的左侧
\vspace{-0.30in}
\centering
\resizebox{\linewidth}{!}{
    \begin{tabular}{c|c|c|c}
     \hline
        Method& L2CD$\times100$ & L1CD$\times100$ & NC \\  % ?????脗搂脙隆?
     \hline
        CVX& 0.020& \textbf{1.038} & 0.906\\
        SIF &0.050&1.600&0.913\\
        Nglod&0.019&1.111&0.934\\
        \hline
        Ours&\textbf{0.018}&\textbf{1.038}&\textbf{0.950}\\
     \hline
   \end{tabular}}
   \vspace{-0.15in}
   \caption{L2CD, L1CD, and NC comparison under Famous.}
   \vspace{-0.3in}
   \label{table:t10DFAUST}
\end{wraptable}

\noindent\textbf{FAMOUS. }We further evaluate our method using non-rigid shapes under FAMOUS dataset. Besides evaluating our surface reconstruction accuracy, we also evaluate our part representation ability. Therefore, we compare our method which also have the ability of representing parts, such as CVX~\cite{DBLP:conf/cvpr/DengGYBHT20} and SIF~\cite{Genova:2019:LST}. Moreover, we also highlight the advantage of surface codes over the method leveraging voxel grids to cover local regions, such as Nglod~\cite{takikawa2021nglod}.

%\begin{table}[tb]
%\centering
%\resizebox{0.7\linewidth}{!}{
%    \begin{tabular}{c|c|c|c}
%     \hline
%        Method& L2CD$\times100$ & L1CD$\times100$ & NC \\  % ?????脗搂脙隆?
%     \hline
%        CVX& 0.020& \textbf{1.038} & 0.906\\
%        SIF &0.050&1.600&0.913\\
%        Nglod&0.019&1.111&0.934\\
%        \hline
%        Ours&\textbf{0.018}&\textbf{1.038}&\textbf{0.950}\\
%     \hline
%   \end{tabular}}
%   \vspace{-0.15in}
%   \caption{L2CD, L1CD, and NC comparison under Famous.}
%   \vspace{-0.23in}
%   \label{table:t10DFAUST}
%\end{table}

For fair comparisons, we train all compared methods to overfit each point cloud in the dataset using the same signed distance supervision. We train our method to regress signed distances by minimizing an MSE loss. We also keep the part number the same in all compared numbers, where we leverage $I=100$ regions to reconstruct surfaces. We employ $125$ latent codes in the $5^3$ voxel grid to produce the results of Nglod. We leverage Euclidean distance $d_E$ to calculate the affinity vector to $I=100$ region centers that are randomly sampled on each point cloud.

\begin{wrapfigure}{r}{0.5\linewidth}% 驴驴脦脛脳脰脛脷脠脻碌脛脳贸虏脿
\vspace{-0.35in}
\includegraphics[width=\linewidth]{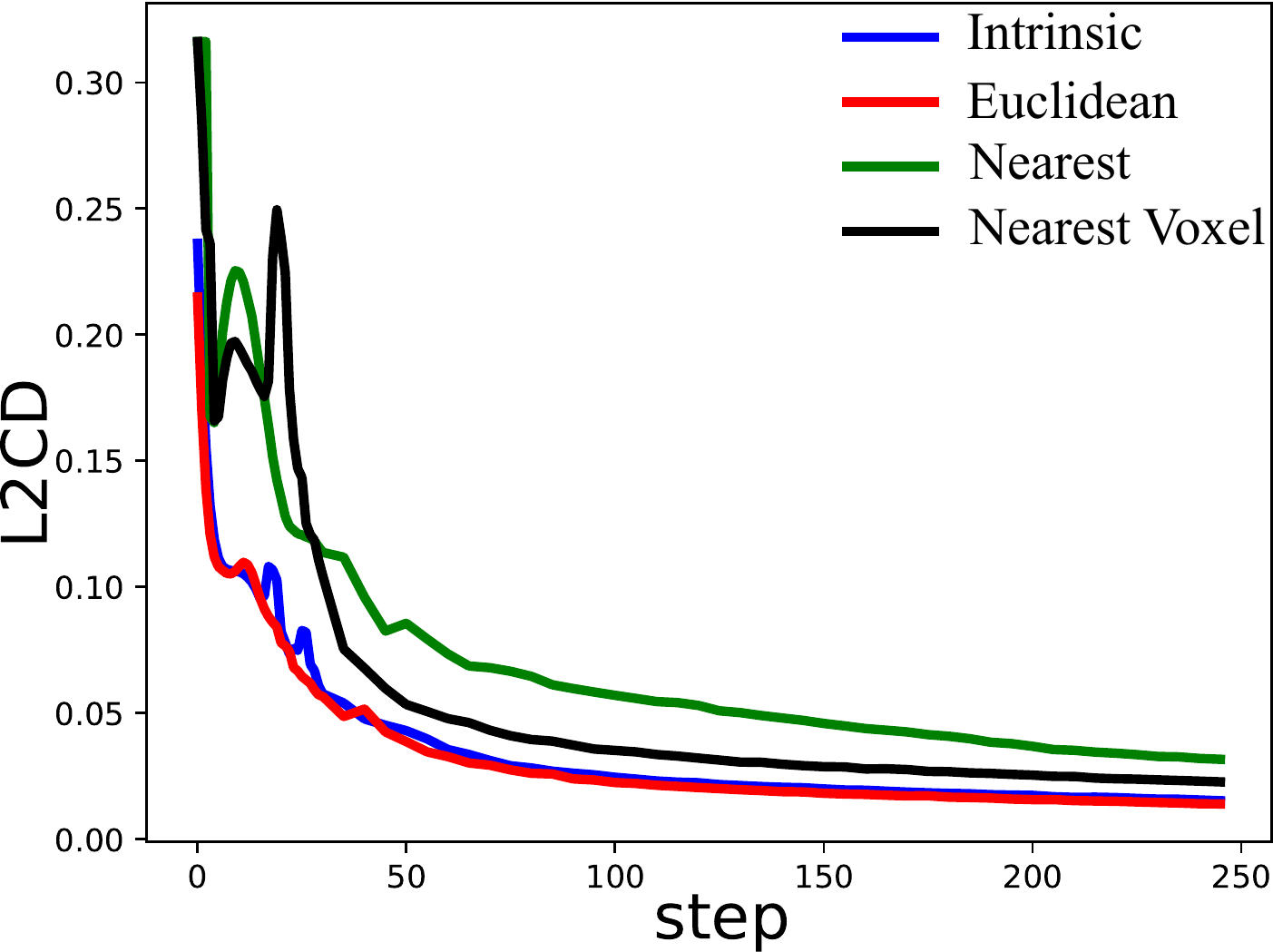}
\vspace{-0.3in}
\caption{\label{fig:CodeComp}Effect of surface codes on training.}
\vspace{-0.34in}
\end{wrapfigure}

\begin{figure*}[tb]
  \centering
  % the following command controls the width of the embedded PS file
  % (relative to the width of the current column)
  %\includegraphics[width=.95\linewidth, bb=39 696 126 756]{figures/definition3.eps}
   \includegraphics[width=\linewidth]{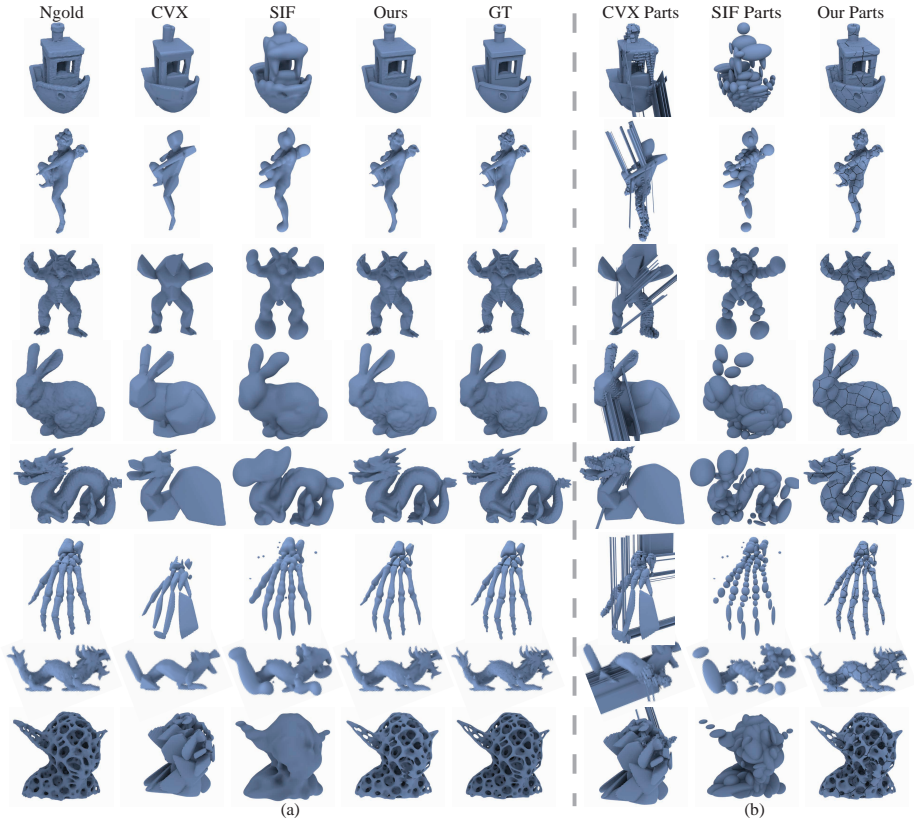}
  % replacing the above command with the one below will explicitly set
  % the bounding box of the PS figure to the rectangle (xl,yl),(xh,yh).
  % It will also prevent LaTeX from reading the PS file to determine
  % the bounding box (i.e., it will speed up the compilation process)
  % \includegraphics[width=.95\linewidth, bb=39 696 126 756]{sampleFig}
  %
  %
  \vspace{-0.3in}
\caption{\label{fig:famous}Visual comparison with Nglod~\cite{takikawa2021nglod}, CVX~\cite{DBLP:conf/cvpr/DengGYBHT20}, SIF~\cite{Genova:2019:LST} in surface reconstruction under FAMOUS in (a). The part comparison is in (b).}
\vspace{-0.34in}
\end{figure*}

We report our numerical comparison in Tab.~\ref{table:t10DFAUST}. We achieve the best performance in terms of all metrics. We further highlight our advantage in visual comparison in Fig.~\ref{fig:famous} (a). We found that our method can reconstruct smoother surfaces with more geometry details than other methods. We also compare the parts that our method reconstructs in Fig.~\ref{fig:famous} (b). Our parts are more meaningful and expressive than others.

\noindent\textbf{D-FAUST. }Finally, we evaluate our method in surface reconstruction for non-rigid shapes. To evaluate our ability of representing articulated parts, we leverage intrinsic distance $d_G$ to calculate the affinity vector. We compare our method with NeuralParts~\cite{DBLP:conf/cvpr/PaschalidouK0F21} which is the latest method for representing articulated shapes. Both of our method and NeuralParts represent a human using $6$ parts. We annotate $I=6$ region centers on the input point cloud. For fair comparisons, we leverage the same signed distance supervision to overfit a shape using our method and NeuralParts.

\begin{table}[tb]
\centering
%\resizebox{\linewidth}{!}{
    \begin{tabular}{c|c|c|c|c}
     \hline
        Class& L2CD$\times100$ & L1CD$\times100$ & NC & IoU\\  % ?????脗搂脙隆?
     \hline
        NeuralParts& 0.008& 0.667 & 0.906 & 0.695\\
        Ours(Parts) &0.007&0.649&0.886 & 0.828\\
        Ours(Shape) &\textbf{0.004} & \textbf{0.555}& \textbf{0.952} & \textbf{0.837}\\
     \hline
   \end{tabular}%}
   %\vspace{-0.15in}
   \caption{L2CD, L1CD, NC and IoU comparison under 10 models of D-FAUST.}
   \vspace{-0.35in}
   \label{table:t10DFaust10}
\end{table}

%\begin{table}[tb]
%\centering
%%\resizebox{\linewidth}{!}{
%    \begin{tabular}{c|c|c|c}
%     \hline
%        Class& L2CD$\times100$ & L1CD$\times100$ & NC\\
%     \hline
%        DeepLS& 0.0065& 0.704 & 0.955\\
%        SIF& 0.0067& 0.681 & 0.945\\
%        Nglod& 0.0062& 0.718 & \textbf{0.958}\\
%        Ours(Euclidean) &0.0064&0.698&0.955\\
%        Ours(Intrinsic) &\textbf{0.0059} & \textbf{0.611}& 0.953\\
%     \hline
%   \end{tabular}%}
%   %\vspace{-0.15in}
%   \caption{L2CD, L1CD and NC comparison under 100 models of D-FAUST.}
%   \vspace{-0.45in}
%   \label{table:t10DFAUST100}
%\end{table}

\begin{wraptable}{r}{0.6\linewidth}% 靠文字内容的左侧
\vspace{-0.50in}
\centering
\resizebox{\linewidth}{!}{
    \begin{tabular}{c|c|c|c}
     \hline
        Class& L2CD$\times100$ & L1CD$\times100$ & NC\\
     \hline
        DeepLS& 0.0065& 0.704 & 0.955\\
        SIF& 0.0067& 0.681 & 0.945\\
        Nglod& 0.0062& 0.718 & \textbf{0.958}\\
        Ours(Euclidean) &0.0064&0.698&0.955\\
        Ours(Intrinsic) &\textbf{0.0059} & \textbf{0.611}& 0.953\\
     \hline
   \end{tabular}}
   \vspace{-0.15in}
   \caption{L2CD, L1CD and NC comparison under 100 models of D-FAUST.}
   \vspace{-0.35in}
   \label{table:t10DFAUST100}
\end{wraptable}

\begin{figure}[tb]
  \centering
  % the following command controls the width of the embedded PS file
  % (relative to the width of the current column)
  %\includegraphics[width=.95\linewidth, bb=39 696 126 756]{figures/definition3.eps}
   \includegraphics[width=\linewidth]{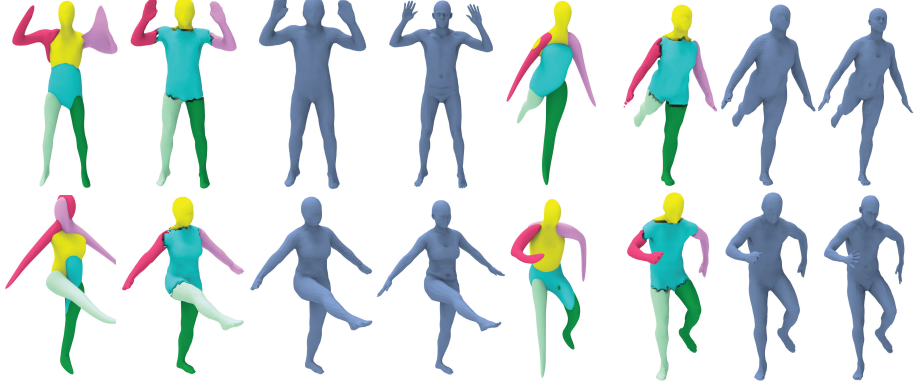}
  % replacing the above command with the one below will explicitly set
  % the bounding box of the PS figure to the rectangle (xl,yl),(xh,yh).
  % It will also prevent LaTeX from reading the PS file to determine
  % the bounding box (i.e., it will speed up the compilation process)
  % \includegraphics[width=.95\linewidth, bb=39 696 126 756]{sampleFig}
  %
  %
  \vspace{-0.35in}
\caption{\label{fig:Human10}Visual comparison with NeuralParts~\cite{DBLP:conf/cvpr/PaschalidouK0F21} under D-FAUST. Color in the same column indicates the same part label.}
\vspace{-0.33in}
\end{figure}

Tab.~\ref{table:t10DFaust10} reports numerical comparison with NeuralParts. We report our results of global reconstruction and local part approximation, both of which outperform the results of NeuralParts. We further demonstrate our performance in visual comparison in Fig.~\ref{fig:Human10}, where our method can reconstruct parts with more geometry details.

\begin{figure}[tb]
\vspace{-0.1in}
  \centering
  % the following command controls the width of the embedded PS file
  % (relative to the width of the current column)
  %\includegraphics[width=.95\linewidth, bb=39 696 126 756]{figures/definition3.eps}
   \includegraphics[width=\linewidth]{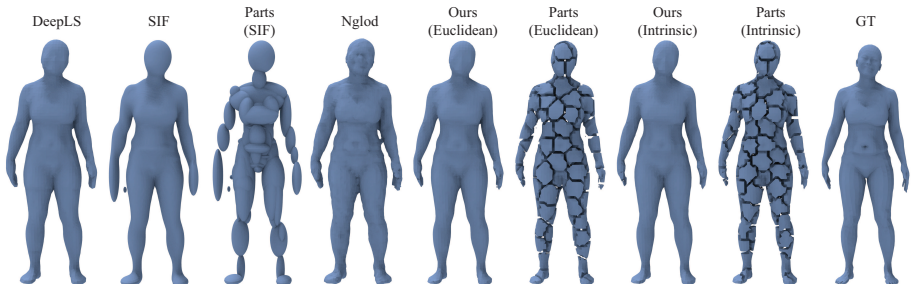}
  % replacing the above command with the one below will explicitly set
  % the bounding box of the PS figure to the rectangle (xl,yl),(xh,yh).
  % It will also prevent LaTeX from reading the PS file to determine
  % the bounding box (i.e., it will speed up the compilation process)
  % \includegraphics[width=.95\linewidth, bb=39 696 126 756]{sampleFig}
  %
  %
  \vspace{-0.33in}
\caption{\label{fig:Human100}Visual comparison with DeepLS~\cite{DBLP:conf/eccv/ChabraLISSLN20}, SIF~\cite{Genova:2019:LST} and Nglod~\cite{takikawa2021nglod} under D-FAUST.}
\vspace{-0.35in}
\end{figure}

We further highlight the advantages of our surface codes in surface reconstruction for humans. We compare our results obtained with Euclidean distances $d_E$ and intrinsic distances $d_G$ with the methods leveraging latent codes on voxel grids or 3D Gaussian functions, including DeepLS~\cite{DBLP:conf/eccv/ChabraLISSLN20}, Nglod~\cite{takikawa2021nglod} and SIF~\cite{Genova:2019:LST}. Similarly, we also produce the result of each method using the same set of signed distance supervision. The set contains $100$ humans, and we train each method to overfit these humans. We leverage $I=100$ latent codes to report our results and the results of SIF, while leveraging $125$ latent codes to report the results of DeepLS and Nglod.

The numerical comparison is shown in Tab.~\ref{table:t10DFAUST100}. Our method achieves the best in terms of CD, and the intrinsic distances $d_G$ work better than Euclidean distances $d_E$. We visualize our reconstruction in Fig.~\ref{fig:Human100}, and our results reveal much smoother surfaces. \vspace{-0.2in}

\subsection{Shape Abstraction}
\begin{figure*}[tb]
  \centering
  % the following command controls the width of the embedded PS file
  % (relative to the width of the current column)
  %\includegraphics[width=.95\linewidth, bb=39 696 126 756]{figures/definition3.eps}
   \includegraphics[width=\linewidth]{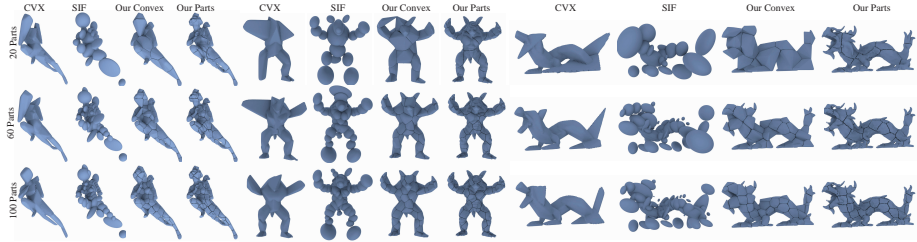}
  % replacing the above command with the one below will explicitly set
  % the bounding box of the PS figure to the rectangle (xl,yl),(xh,yh).
  % It will also prevent LaTeX from reading the PS file to determine
  % the bounding box (i.e., it will speed up the compilation process)
  % \includegraphics[width=.95\linewidth, bb=39 696 126 756]{sampleFig}
  %
  %
  \vspace{-0.3in}
\caption{\label{fig:famousabstraction}Visual comparison with CVX~\cite{DBLP:conf/cvpr/DengGYBHT20} and SIF~\cite{Genova:2019:LST} in shape abstraction with different numbers of parts.}
\vspace{-0.30in}
\end{figure*}

Our method can also represent a shape as an abstraction. We remove the geometry details on a shape by representing each one of its parts as a convex hull. We leverage our learned model in Tab.~\ref{table:t10DFAUST} to produce the abstractions for shapes in FAMOUS dataset. CVX and SIF also produced results in an overfitting way as ours. Visual comparison in Fig.~\ref{fig:famousabstraction} demonstrates that our method can reveal more complex structures on shapes with the same number of parts.

%\begin{figure}[tb]
%  \centering
%  % the following command controls the width of the embedded PS file
%  % (relative to the width of the current column)
%  %\includegraphics[width=.95\linewidth, bb=39 696 126 756]{figures/definition3.eps}
%   \includegraphics[width=\linewidth]{SegmentationInstance-eps-converted-to.pdf}
%  % replacing the above command with the one below will explicitly set
%  % the bounding box of the PS figure to the rectangle (xl,yl),(xh,yh).
%  % It will also prevent LaTeX from reading the PS file to determine
%  % the bounding box (i.e., it will speed up the compilation process)
%  % \includegraphics[width=.95\linewidth, bb=39 696 126 756]{sampleFig}
%  %
%  %
%  \vspace{-0.3in}
%\caption{\label{fig:instanceseg}Shape abstraction with instance segmentation. Color in the same row indicates the same part label.}
%\vspace{-0.31in}
%\end{figure}

\begin{wrapfigure}{r}{0.6\linewidth}% 驴驴脦脛脳脰脛脷脠脻碌脛脳贸虏脿
\vspace{-0.2in}
\includegraphics[width=\linewidth]{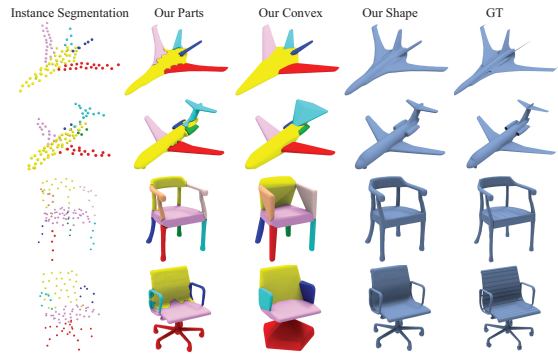}
\vspace{-0.3in}
\caption{\label{fig:instanceseg}Shape abstraction with instance segmentation. Color in the same row indicates the same part label.}
\vspace{-0.35in}
\end{wrapfigure}

With semantic distance $d_S$ to obtain affinity vector, our method can also produce shape abstraction with more meaningful parts. Besides the ability of encoding semantic segmentation in Fig.~\ref{fig:segmentation}, we report our results learned by Eq.~(\ref{eq:5}) with instance segmentation (100 points) in Fig.~\ref{fig:instanceseg}. We first reconstruct the shape and the parts, and abstract the shape using convex hulls of its parts. Due to our ability of using the surface attributes, our method not only reconstructs smooth surfaces but also reveals plausible parts. More abstraction results are in our supplemental materials.\vspace{-0.2in}

\subsection{Ablation Study}
\noindent\textbf{Distance Metrics. }We compare the effect of distance metrics on the performance under the $10$ humans from D-FAUST. We compare the results with Euclidean distance $d_E$, intrinsic distance $d_G$, and no distance encoding. Without distance encoding, we produce the affinity vector as a uniform vector (``Average'') or a one-hot vector indicating the nearest surface code (``Nearest''). The numerical comparison in Tab.~\ref{table:distancemetric} demonstrates that blending with either Euclidean distances or intrinsic distances achieves better performance and intrinsic distances are the best for non-rigid shape modeling.

%\begin{table}[tb]
%\centering
%%\resizebox{\linewidth}{!}{
%    \begin{tabular}{c|c|c|c}
%     \hline
%        Affinity& L2CD$\times$100 & L1CD$\times$100 & NC\\
%     \hline
%        Average &0.0058&0.671&\textbf{0.952}\\
%        Nearest&0.0060 & 0.671& 0.950\\
%        Euclidean & 0.0057& 0.661 & \textbf{0.952}\\
%        Intrinsic & \textbf{0.0039}& \textbf{0.555} & \textbf{0.952}\\
%     \hline
%   \end{tabular}%}
%   \vspace{-0.15in}
%   \caption{Ablation studies under 10 models of D-FAUST.}
%   \vspace{-0.195in}
%   \label{table:distancemetric}
%\end{table}

\begin{wraptable}{r}{0.5\linewidth}% 靠文字内容的左侧
\vspace{-0.05in}
\centering
%\resizebox{\linewidth}{!}{
    \begin{tabular}{c|c|c|c}
     \hline
        Affinity& L2CD$\times$100 & L1CD$\times$100 & NC\\
     \hline
        Average &0.0058&0.671&\textbf{0.952}\\
        Nearest&0.0060 & 0.671& 0.950\\
        Euclidean & 0.0057& 0.661 & \textbf{0.952}\\
        Intrinsic & \textbf{0.0039}& \textbf{0.555} & \textbf{0.952}\\
     \hline
   \end{tabular}%}
   \vspace{-0.1in}
   \caption{Ablation studies under 10 models of D-FAUST.}
   \vspace{-0.35in}
   \label{table:distancemetric}
\end{wraptable}

\noindent\textbf{Convergence. }We compare the effect of surface codes on convergence under the $10$ humans from D-FAUST. We compare the average L2CD obtained with Euclidean distance $d_E$, intrinsic distance $d_G$, one-hot affinity vector (``Nearest'') and latent codes on voxel vertices (``Nearest Voxel''). The comparison in Fig.~\ref{fig:CodeComp} indicates that blending parts with $d_E$ or $d_G$ makes the training converge faster and better than only using the nearest code on both surface or voxel grids.

\begin{wrapfigure}{r}{0.6\linewidth}% 驴驴脦脛脳脰脛脷脠脻碌脛脳贸虏脿
\vspace{-0.2in}
\includegraphics[width=\linewidth]{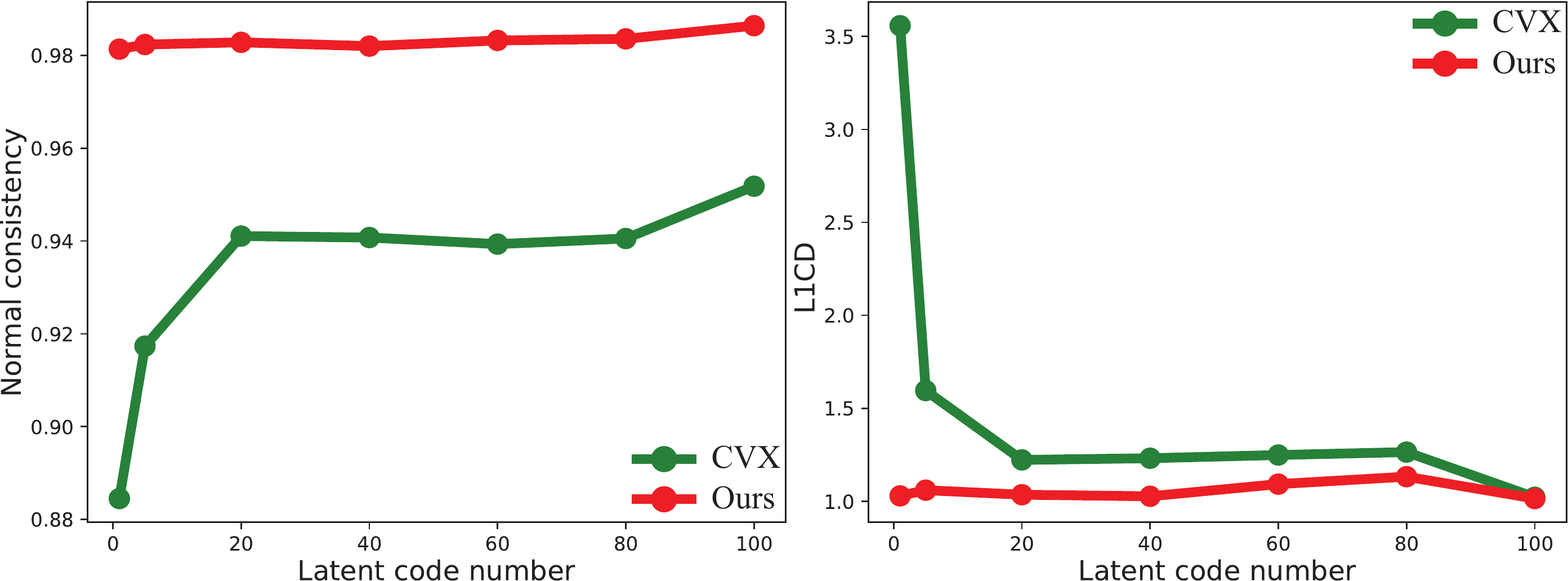}
\vspace{-0.3in}
\caption{\label{fig:CodeNumber}Effect of surface code number.}
\vspace{-0.35in}
\end{wrapfigure}

\noindent\textbf{Surface Code Number. }Another advantage of surface codes is that we can represent shapes at multilevel without significant geometry details loss. We reconstruct the Stanford bunny using different numbers of surface codes. Fig.~\ref{fig:CodeNumber} indicates that our method still achieves high accuracy even with few surface codes, while CVX~\cite{DBLP:conf/cvpr/DengGYBHT20} requires more parts to approximate the shape well. The reason is that our method represents parts and blends parts in the latent space, which achieves better representation ability.\vspace{-0.2in}

%\begin{figure}[tb]
%  \centering
%  % the following command controls the width of the embedded PS file
%  % (relative to the width of the current column)
%  %\includegraphics[width=.95\linewidth, bb=39 696 126 756]{figures/definition3.eps}
%   \includegraphics[width=0.8\linewidth]{CodeNumber-eps-converted-to.pdf}
%  % replacing the above command with the one below will explicitly set
%  % the bounding box of the PS figure to the rectangle (xl,yl),(xh,yh).
%  % It will also prevent LaTeX from reading the PS file to determine
%  % the bounding box (i.e., it will speed up the compilation process)
%  % \includegraphics[width=.95\linewidth, bb=39 696 126 756]{sampleFig}
%  %
%  %
%  \vspace{-0.12in}
%\caption{\label{fig:CodeNumber}Effect of surface code number.}
%\vspace{-0.3in}
%\end{figure}

%\begin{table}[tb]
%\centering
%\resizebox{\linewidth}{!}{
%    \begin{tabular}{c|c|c|c|c}
%     \hline
%        method & center number& CD2\times100 & CD\times100 & normal\\
%     \hline
%	\multirow{cvxnet} & 1 & 0.4499 & 3.558 & 0.884 \\
%	 & 5 & 0.0504 & 1.595 & 0.917\\
%	 & 20 & 0.0228 & 1.222 & 0.941\\
%	 & 40 & 0.0234 & 1.231 & 0.941\\
%	 & 60 & 0.0229 & 1.248 & 0.939\\
%	 & 80 & 0.0238 & 1.264 & 0.941\\
%	 & 100 & 0.0130 & 1.021 & 0.952\\ \hline
%	 \multirow{ours} & 1 & 0.0121 & 1.029 & 0.981 \\
%	 & 5 & 0.0126 & 1.059 & 0.982 \\
%	 & 20 & 0.0121 & 1.035 & 0.983\\
%	 & 40 & 0.0119 & 1.026 & 0.982\\
%	 & 60 & 0.0132 & 1.092 & 0.983\\
%	 & 80 & 0.0141 & 1.132 & 0.984\\
%	 & 100 & \textbf{0.0112} & \textbf{1.014} & \textbf{0.986} \\ \hline
%
%   \end{tabular}}
%   \caption{Center number effect under bunny models of FAMOUS.}
%   \label{table:t10}
%\end{table}

%\noindent\textbf{Limitation. }LPI can not adjust the part centers, which makes it unable to split a shape adaptively.

\section{Conclusion}
We introduce latent partition implicit to represent 3D shapes. LPI is a multi-level representation, which efficiently represents a shape using different numbers of parts. Our method successfully represents parts using surface codes, and blend parts by weighting surface codes in the latent space to reconstruct surfaces. This leads to highly accurate shape and part modeling. With surface codes, latent partition implicit also enables to flexibly combine additional surface attributes, such as geodesic distance or segmentation. We can learn latent partition implicit from point clouds without requiring ground truth signed distances or point normals. Our method outperforms the state-of-the-art under widely used benchmarks.

\clearpage
% ---- Bibliography ----
%
% BibTeX users should specify bibliography style 'splncs04'.
% References will then be sorted and formatted in the correct style.
%
\bibliographystyle{splncs04}
\bibliography{papers}
\end{document}